\documentclass{article}

\usepackage{PRIMEarxiv}

\usepackage{algorithm}
\usepackage{algpseudocode}
\usepackage{caption}
\usepackage{subcaption}

\usepackage[utf8]{inputenc} 
\usepackage[T1]{fontenc}    
\usepackage{hyperref}       
\usepackage{url}            
\usepackage{booktabs}       
\usepackage{amsfonts}       
\usepackage{nicefrac}       
\usepackage{microtype}      
\usepackage{lipsum}
\usepackage{fancyhdr}       
\usepackage{graphicx}       
\graphicspath{{media/}}     

\title{Multilingual Information Retrieval with a Monolingual Knowledge Base}

\author{Yingying Zhuang\\
  Amazon \\
  {\tt yyzhuang@amazon.com} \\\And
  Aman Gupta\\
  Amazon \\
  {\tt amangta@amazon.com} \\\And
  Anurag Beniwal \\
  Amazon \\
  {\tt beanurag@amazon.com} \\}

\begin{document}
\maketitle

\begin{abstract}
Multilingual information retrieval has emerged as powerful tools for expanding knowledge sharing across languages. On the other hand, resources on high quality knowledge base are often scarce and in limited languages, therefore an effective embedding model to transform sentences from different languages into a feature vector space same as the knowledge base language becomes the key ingredient for cross language knowledge sharing, especially to transfer knowledge available in high-resource languages to low-resource ones. In this paper we propose a novel strategy to fine-tune multilingual embedding models with weighted sampling for contrastive learning, enabling multilingual information retrieval with a monolingual knowledge base. We demonstrate that the weighted sampling strategy produces performance gains compared to standard ones by up to 31.03\% in MRR and up to 33.98\% in Recall@3. Additionally, our proposed methodology is language agnostic and applicable for both multilingual and code switching use cases. 
\end{abstract}


\section{Introduction}
The task of Information Retrieval is to find relevant information from a large collection and a knowledge base often complements this retrieval task and facilitates various downstream tasks as well. With the recent emergence of Large Language Models (LLMs) as a powerful tool to build dialogue systems \cite{DialogSumm, zhuang-etal-2021-weakly,zhang2025reicragenhancedintentclassification} and conversation AIs, the retrieved information can play a critical role in increasing practicality and reliability of these systems. Application of LLMs has also achieved remarkable advancements in multilingual scenarios and many of the current frontier LLMs, such as Anthropic’s Claude 3 \cite{claude3} and Cohere's Command A \cite{cohere2025commandaenterprisereadylarge} are focused to excel in the multilingual settings to support global businesses. On the other hand, because knowledge base construction is labor-intensive and expensive \cite{paulheim2018much}, imbalanced data, quality and scarcity issues become the bottleneck for rapid knowledge base development, especially for low-resource languages. Additionally there has been a surge of interest in code-switching recently where communicators switch between two or more languages during linguistic interactions \cite{dogruoz-etal-2021-survey, winata-etal-2021-multilingual}. This communication style is common in bilingual and multilingual communities \cite{zhang2023speakforeignlanguagesvoice, huang2024zeroresourcecodeswitchedspeech}. For example, in US mixing Spanish and English is a common phenomenon while in Canada mixing French and English has been often observed. However, it remains a challenging task in Information Retrieval as the language IDs are not specified and knowledge base with code-switching context is scarce. Existing work has been mainly focused on automatic multilingual knowledge base construction \cite{Ji_2022,zhou2021prix} where knowledge bases of low-resource languages are automatically propagated based on knowledge bases in well-populated high resource languages. 

\paragraph{Contributions} Different from previous approaches, we propose a novel embedding model development pipeline to align multilingual information retrieval with a \textbf{monolingual} knowledge base, therefore removing the bottleneck of multilingual knowledge base construction in information retrieval systems. Specifically, we design a weighted sampling strategy to select positive and negative pairs for contrastive learning, which guides the model to embed queries with similar semantic meanings into close embedding vector space across different languages. Our approach aims at a more data-efficient multilingual information retrieval system which only requires a high resource language (e.g., English or French) for the knowledge base. Experimental results demonstrate that our approach enables an embedding model to properly handle multilingual queries. The ability to share knowledge in multiple languages is a fundamental component to ensure the development of Conversation AIs systems and to enable both businesses and individuals to reach a broader range of communities, fostering inclusivity, accessibility and collaboration among international teams. 

\section{Methodology}
Embedding models such as E5 \cite{Wang2022TextEB}, GTE \cite{Li2023TowardsGT}, Nomic \cite{nussbaum2025nomicembedtrainingreproducible}, and Arctic-Embed \cite{merrick2024arcticembedscalableefficientaccurate} are typically trained with three stages: pretraining via masked language modeling, large scale contrastive pretraining with in-batch negatives, and finally quality focused contrastive fine-tuning \cite{ wang2024multilinguale5textembeddings,chen2024bgem3embeddingmultilingualmultifunctionality,zhang2024mgtegeneralizedlongcontexttext}. Our work focuses on the contrastive fine-tuning stage and the proposed methodology can be applied on any open-source text embedding models. One key ingredient for effective embedding is supervised fine-tuning with a high quality dataset which contains explicitly labeled positive and negative pairs. Specifically, we focus on the query-to-query multilingual retrieval problem with a monolingual knowledge base. 

Given a user query $q$ in a target language $T$, the goal is to retrieve a similar example query $q_{\textsc{ex}}$ from the knowledge base in language $E$ where $E$ is different from $T$. The knowledge base contains a set of example queries and their corresponding labels $$KB_E = {(q_{\textsc{ex}_1}, l_1), (q_{\textsc{ex}_2}, l_2),..., (q_{\textsc{ex}_N}, l_N) }.$$ For example, a key component of task-oriented dialogue systems is to identify the underlying purpose or intent of a user in a conversation \cite{zhang2025reicragenhancedintentclassification,degand-muller-2020-introduction}. To facilitate information retrieval business can build a monolingual intent knowledge base in English which contains a set of customer query and customer intent pairs based on high quality data source or human annotation. As business expands to other languages, it becomes infeasible to build a separate knowledge base in each language therefore the goal is to share the English knowledge base across all languages. If a user input query is in Spanish (the target language $T$), the goal is to retrieve a semantically similar English query in the knowledge base, $q_{\textsc{ex}_j}$, and the corresponding intent label $l_j$. In order to build a multilingual Information Retrieval system with an English-only (or any other monolingual) knowledge base, it is crucial for the embedding model to map English and non-English queries into a shared representation space to enable non-English retrieval. 

\paragraph{Contrastive Training Data} In contrastive learning, the model learns by distinguishing between similar and dissimilar queries from positive and negative pairs. An effective embedding model will project similar (positive) pairs closer together and dissimilar (negative) pairs further apart in a embedding vector space. For positive pairs, existing work often uses a noisy data source to identify relevant queries, then applies some heuristic and consistency quality filters to improve data quality \cite{nussbaum2025nomicembedtrainingreproducible, günther2023jinaembeddingsnovelset,wang-etal-2024-improving-text, dai2022promptagatorfewshotdenseretrieval}. For negative pairs, \cite{qu2021rocketqaoptimizedtrainingapproach} demonstrated the importance of training on hard negatives, where “hard” refers to the fact that it is not trivial to determine their lower relevance relative to the positive examples, therefore, many existing work has been following this paradigm to identify the hardest negatives for each training example. For example, \cite{merrick2024arcticembedscalableefficientaccurate} leveraged a pre-existing text embedding model to identify and score the hardest negatives for each training example while NV Retriever \cite{moreira2025nvretrieverimprovingtextembedding} also added a filtering step where any negative with a relevance score exceeding a specified percentage of the known-positive’s score is discarded as a potential false negative. 

\begin{figure*}[t]
\centering
\includegraphics[width=\textwidth]{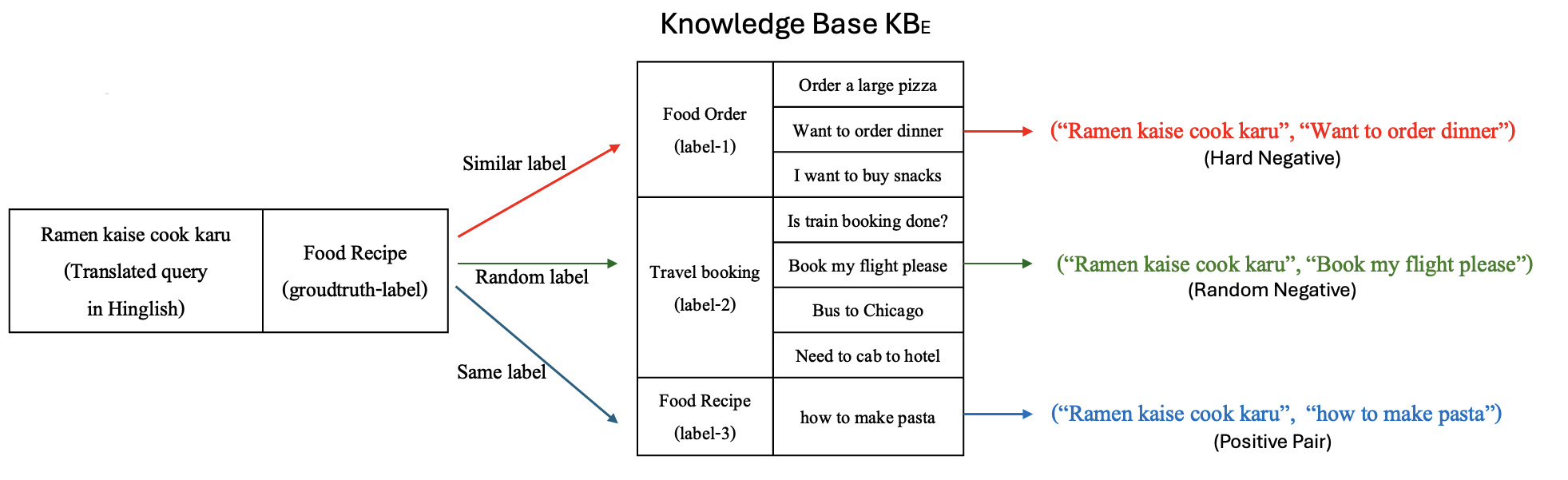}
\caption{Illustration of our contrastive training data generation algorithm. For each translated query (left), we create a positive pair (blue) with queries sharing the same label, k hard negative pairs (red) from queries with similar labels selected through weighted sampling, and 1 random negative pair (green) from queries with different labels selected randomly from the knowledge base.}
\label{fig:figfinal}
\end{figure*}

\paragraph{Contrastive Training Data Generation}
In our methodology, we propose a new strategy to leverage the available labels, $l_1, l_2, ..., l_n$, in the monolingual knowledge base to construct positive and negative multilingual pairs. Our approach demonstrates the benefits of a weighted sampling strategy of negative mining based on relevance, which in Section~\ref{sec:exp} we show that it is more effective than focusing purely on selecting the "hardest" negatives possible, or a rank for relevance scores. Algorithm \ref{alg1} shows the details of our proposed approach step-by-step and Figure ~\ref{fig:figfinal} is an illustration with examples. 
\begin{algorithm}
\caption{Contrastive Training Data Generation}\label{alg1}
\begin{algorithmic}
\Statex \textbullet~\textbf{Require:} 
\State A knowledge base in language $E$ with $N$ example query and label pairs $$KB_E = {(q_{\textsc{ex}_1}, l_1), (q_{\textsc{ex}_2}, l_2),..., (q_{\textsc{ex}_N}, l_N) }.$$
\State A set of $M$ unlabelled queries $q$'s in a target language $T$ $(q_1, q_2, ..., q_M)$.

\Statex \textbullet~\textbf{Data Generation} 
\State \textopenbullet~\textbf{Step 1}: \textbf{Initialize} an empty paired dataset $P = \{\} $
\State \textopenbullet~\textbf{Step 2}: \textbf{Split} $KB_E$ into index and training set. We split the $N$ example query and label pairs into two subsets $N_1$ and $N_2$, reserve the $N_1$ pairs for retrieval index, while using the remaining $N_2$ for systhetic data generation.
\State \textopenbullet~\textbf{Step 3}: \textbf{Generate Positive Pairs}: For each query $q_i$ in $N_2$ which is in language $E$, use an LLM to translate it to the target language $T$ into $q_i^{T}$. Randomly sample one query from all the queries sharing the same label in the Index set $N_1$, $q_i^{'}$. Update $P$ with the positive pair $(q_i^{T}, q_i^{'})$, i.e. $P = P\cup{(q_i^{T}, q_i^{'})}$.
\State \textopenbullet~\textbf{Step 4}: \textbf{Generate Negative Pairs based on weighted sampling}: For each query $q_i$ in $N_2$, use an LLM to translate it to the target language $T$ into $q_i^{T}$ (or reuse $q_i^{T}$ from Step 3). For each query $q_j$ in the index set $N_1$, calculate a similarity score $s_{ij}$ between $q_i$'s label $l_i$ and $q_j$'s label $l_j$. We generate two types of negative pairs:
\begin{itemize}
    \item \textbf{Random negative}: Randomly sample one query $q_i^{''}$ from all the queries in set $N_1$ with a different label than $l_i$.
    \item \textbf{Hard negatives}: Randomly sample $k$ queries from the index set $N_1$ based on sampling weight $s_{ij}$. The higher the similarity score $s_{ij}$, the more likely a query is to be sampled. The intuition is that queries with similar labels are harder to distinguish, making them better candidates for hard negatives.
\end{itemize}
With $k=2$, we yield three negative pairs in total: one random negative pair $(q_i^{T}, q_i^{''})$ and two hard negative pairs $(q_{i}^{T}, q_i^{'''})$ and $(q_i^{T}, q_i^{''''})$. This maintains our desired positive to negative ratio of 1:3 when using contrastive loss. Update $P$ with these negative pairs, i.e., $P = P\cup\{(q_i^{T}, q_i^{''}), (q_i^{T}, q_i^{'''}), (q_i^{T}, q_i^{''''})\}$.
\State \textopenbullet~\textbf{Step 5}: \textbf{Synthetic Data Augmentation}: Compared to the abundance of unlabeled data, high-quality labeled examples in $KB_E$ are more scarce. To address this data scarcity issue, we also use synthetic data creation to construct additional positive and negative pairs. For any query in the target language $T$, use a LLM to generate a semantically similar query to form a positive pair, and 3 semantically different queries in language $E$ to form 3 negative pairs. Update $P$ with the these pairs.
\end{algorithmic}
\end{algorithm}

\section{Experiments and Results}\label{sec:exp}
In this paper, we conduct experiments to evaluate the plausibility of our proposed methods for the business use case of performing Information Retrieval for Hinglish queries using an English knowledge base. Hinglish refers to the conversation style where speakers are mixing Hindi and English in one conversation interaction and often within one single sentence, which is a very common way of communication for the India marketplace.

Our proposed methodology is language agnostic and can be applied to any language with low-resource constraints. Therefore, we simply treat code-switching as a new language. In our experiments we fine-tune an embedding model optimzed to embed English and Hinglish queries into the same vector space for Information Retrieval.

\subsection{Datasets}
\paragraph{Data Anonymization} Due to business considerations, we are not permitted to share the results using the original customer queries. As a result, we manually anonymized both the labels and transcripts to ensure no personal information is included. Additionally, specific product and service names were denonymized to prevent the identification of the company from the transcript or label descriptions. Despite these modifications, the conclusions drawn from our experiments remain valid. 

We collect about 35,000 dialogue transcripts with customer intent labels in English to construct our $KB_E$. Following the steps in Algorithm \ref{alg1} we split these transcripts into two subsets: ~3000 to construct the retrieval index and the remaining 32,000 to generate constrastive training data. For Step 5, we collect 10,000 unlabelled dialogue transcripts in Hinglish for data augmentation. 

\subsection{Experiments}
As discussed earlier, we proposed Algorithm~\ref{alg1}, which employs a hybrid approach combining both hard and random negative mining, along with synthetic data generation. This mixed strategy helps reduce excessive reliance on translated queries alone, with synthetic data generation performed directly on Hinglish data available natively.

In our experimental setup, we first evaluated several open-source multilingual retrieval models as the foundation for our fine-tuning process:
\begin{itemize}
    \item {\texttt{paraphrase-multilingual-mpnet-base-v2}}
    \item {\texttt{multilingual-e5-base}}
    \item {\texttt{paraphrase-multilingual-MiniLM-L12-v2}}
    \item {\texttt{stsb-xlm-r-multilingual}}
\end{itemize}

To validate the effectiveness of our proposed approach and to better understand the contribution of different components, we conducted ablation studies comparing Algorithm~\ref{alg1} against several alternative strategies:

\paragraph{Negative Sampling Variations}
\begin{itemize}
    \item \textbf{Random Negative Mining:} This represents the simplest approach where we set $k=0$ in Step 4 of Algorithm~\ref{alg1}, selecting all negatives randomly with equal probability regardless of semantic similarity.
    
    \item \textbf{Hard Negative Mining:} In this variation, we set $k=3$ in Step 4, selecting all negatives using weighted sampling based on similarity scores. This targets examples that are more challenging to distinguish.
    
    \item \textbf{Hardest Negative Mining:} Taking the hard negative approach further, we not only set $k=3$ but also restrict sampling to only the top-3 most similar labels (excluding exact matches). This focuses exclusively on the most challenging negative examples.
\end{itemize}

\paragraph{Data Source Variations}
To evaluate the impact of synthetic data augmentation described in Step 5, we also tested:
\begin{itemize}
    \item \textbf{Labeled Data Only:} This variant trains solely on the high-quality translated examples from the original English dataset, omitting the synthetic data generation Step 5.
    
    \item \textbf{Synthetic Data Only:} Conversely, this approach relies exclusively on synthetically generated data in Step 5, representing the scenario where no labeled data is available in any language.
\end{itemize}

These methodical comparisons allowed us to isolate the contribution of each component in our algorithmic design and confirm the superiority of our hybrid approach.

\subsection{Implementation Details}
We fine-tuned various pretrained multilingual embedding models using the InfoNCE contrastive loss\cite{Oord2018RepresentationLW} with a maximum sequence length of 512 tokens. Training was performed on 4 NVIDIA A10G GPUs using PyTorch's DistributedDataParallel framework with a batch size of 32. Each model was trained for 15 epochs, with early convergence typically observed around epochs 8-9. For optimization, we used AdamW with a learning rate of 2e-5 and a linear warmup period. Claude Sonnet 3.5\footnote{\url{https://www.anthropic.com/news/claude-3-5-sonnet}} was employed for both query translation and generation of synthetic positive and negative pairs for training.

\subsection{Evaluation Metrics}
For evaluation, we compute Recall@1, Recall@3, Recall@5, and Recall@10 metrics, along with Mean Reciprocal Rank (MRR). These metrics are particularly appropriate for our task since we have a single ground truth label for each query, while our retrieval system returns ranked predictions. Recall@k measures whether the correct label appears within the top-k retrieved results, providing a clear assessment of our model's retrieval performance at various thresholds of interest.

\subsection{Results and Discussion}
\begin{figure}[h]
  \centering
  \includegraphics[scale=0.6]{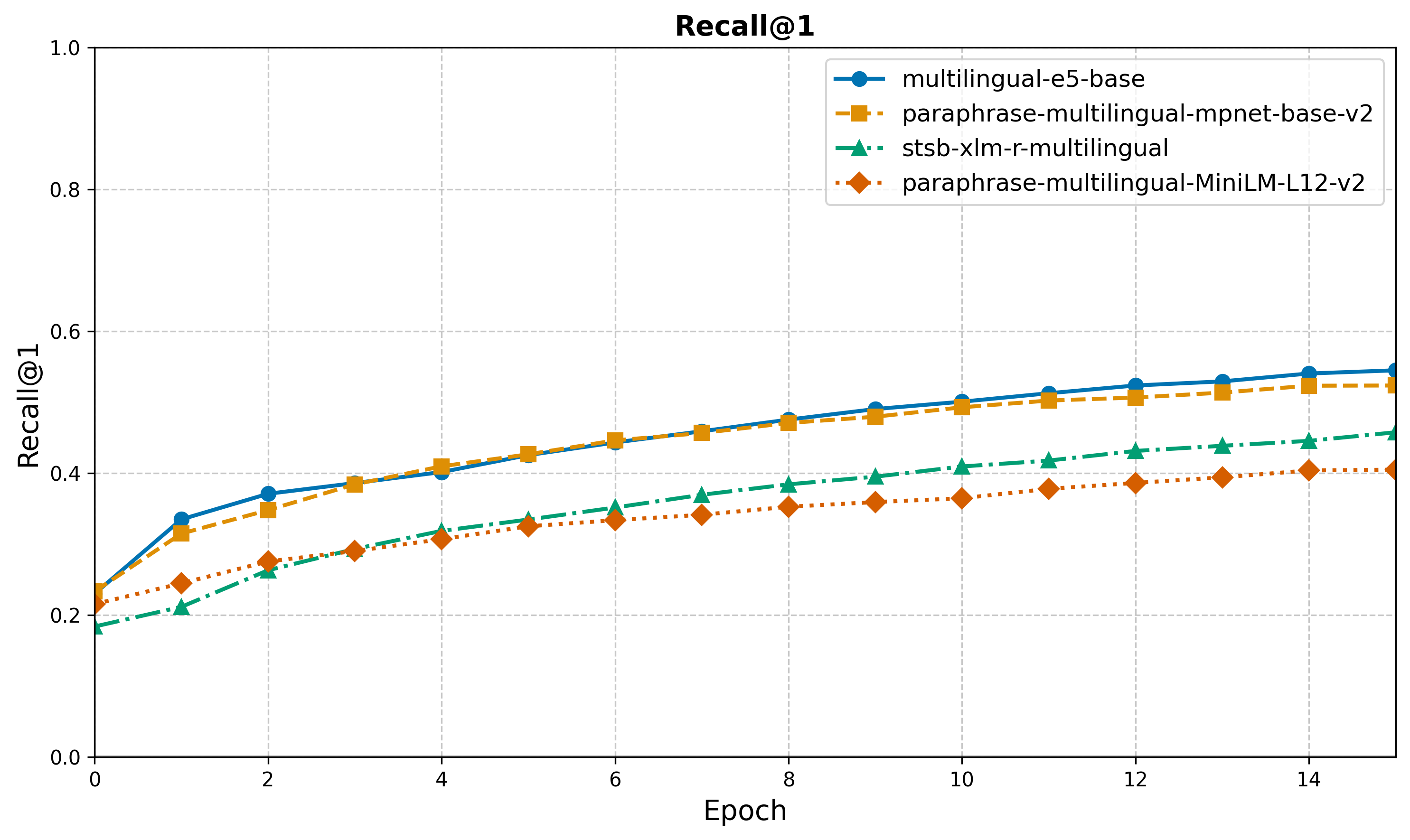}
  \caption{Performance of Recall@1 at different points during contrastive fine-tuning.}
\label{fig:top1}
\end{figure}

Figure~\ref{fig:top1} shows the performance comparison of different multilingual embedding models during training. The \textit{multilingual-e5-base} model consistently outperforms other models across all epochs, reaching the highest Recall@1 of approximately 0.54 by the final epoch. The \textit{paraphrase-multilingual-mpnet-base-v2} model performs comparably well, while \textit{stsb-xlm-r-multilingual} and \textit{paraphrase-multilingual-MiniLM-L12-v2} show significantly lower performance. Based on these results, we selected \textit{multilingual-e5-base} as our foundation model for all subsequent experiments. Appendix Section \ref{oth-plots} presents the comparison plots between these models on Recall@3, Recall@5, Recall@10, and MRR, which show a similar pattern as Recall@1.

The evaluation results for different negative sampling and data strategies are presented in Table~\ref{tab:results}. Our ablation study reveals that combining random and hard negatives yields better performance than approaches using only one type of negative examples. While the Labeled Data Only approach performs comparably to our proposed method, we attribute this to the composition of our test set, which primarily contains samples similar to the labeled data. Nevertheless, Algorithm 1's hybrid approach provides better protection against performance degradation on original queries in the low-resource language. The significantly poorer performance of the purely synthetic data approach indicates the crucial importance of high-quality labeled examples in the training process. 

\textbf{Why does mixed-mining strategy work better?}  The superior performance of Algorithm~\ref{alg1} can be attributed to its hybrid approach. By combining both random and hard negatives, the model optimizes the embedding space globally while simultaneously emphasizing differentiation in local neighborhoods. Pure hard negative mining (0.4613) or hardest negative mining (0.4012) approaches underperform because they overly focus on disambiguating local neighborhoods without maintaining global embedding structure. Additionally, the incorporation of synthetic data in Algorithm~\ref{alg1} helps improve performance on low-resource language queries, reducing overfitting to the translated query distribution while maintaining strong performance on the original dataset.

\begin{table}[ht]
\centering
\caption{Performance comparison of different negative sampling and data strategies}
\label{tab:results}
\begin{tabular}{lcccc}
\toprule
\textbf{Method} & \textbf{Top-1} & \textbf{Top-3} & \textbf{Top-10} & \textbf{MRR} \\
\midrule
Random Negative Mining & 0.5308 & 0.7271 & 0.8794 & 0.6520 \\
Hard Negative Mining & 0.4613 & 0.6829 & 0.8535 & 0.5982 \\
Hardest Negative Mining & 0.4012 & 0.6678 & 0.8499 & 0.5610 \\
Labeled Data Only & \textbf{0.5474} & 0.7356 & 0.8803 & 0.6639 \\
Synthetic Data Only & 0.2191 & 0.4012 & 0.6444 & 0.3550 \\
\midrule
Using Algorithm \ref{alg1} & 0.5450 & \textbf{0.7410} & \textbf{0.8842} & \textbf{0.6653} \\
\bottomrule
\end{tabular}
\end{table}

\section{Conclusion and Future Work}
Our proposed embedding model development pipeline to align multilingual information retrieval with a monolingual knowledge base removes the bottleneck of multilingual knowledge base construction in information retrieval systems. The ability to share knowledge across languages facilitates faster and easier global expansion of conversation AI systems, specifically to lower resource languages or code-switching use cases. Our comprehensive experiment results in a code-switching use case demonstrate the effectiveness and robustness of our proposed weighted sampling strategy for contrastive learning. Our framework is language agnostic therefore applicable for any target language and we hope our paper can help push forward the research in the multilingual information retrieval communities and facilitate faster global expansion for businesses. 

In the future, we seek to continue our experimentation to leverage monolingual knowledge base for multilingual dialogue systems. Additionally, we hope to explore more robust embedding models with compression approaches such as binarization or quantization to further improve performance.

\bibliographystyle{unsrt}  
\bibliography{references}  

\begin{thebibliography}{10}

\bibitem{DialogSumm}
Yingying Zhuang, Jiecheng Song, Narayanan Sadagopan, and Anurag Beniwal.
\newblock Self-supervised pre-training and semi-supervised learning for extractive dialog summarization.
\newblock In {\em Companion Proceedings of the ACM Web Conference 2023}, WWW '23 Companion, page 1069–1076, New York, NY, USA, 2023. Association for Computing Machinery.

\bibitem{zhuang-etal-2021-weakly}
Yingying Zhuang, Yichao Lu, and Simi Wang.
\newblock Weakly supervised extractive summarization with attention.
\newblock In Haizhou Li, Gina-Anne Levow, Zhou Yu, Chitralekha Gupta, Berrak Sisman, Siqi Cai, David Vandyke, Nina Dethlefs, Yan Wu, and Junyi~Jessy Li, editors, {\em Proceedings of the 22nd Annual Meeting of the Special Interest Group on Discourse and Dialogue}, pages 520--529, Singapore and Online, July 2021. Association for Computational Linguistics.

\bibitem{zhang2025reicragenhancedintentclassification}
Ziji Zhang, Michael Yang, Zhiyu Chen, Yingying Zhuang, Shu-Ting Pi, Qun Liu, Rajashekar Maragoud, Vy~Nguyen, and Anurag Beniwal.
\newblock Reic: Rag-enhanced intent classification at scale, 2025.

\bibitem{claude3}
Introducing the next generation of claude.
\newblock \url{https://www.anthropic.com/news/claude-3-family}.
\newblock Accessed: 2025-04-20.

\bibitem{cohere2025commandaenterprisereadylarge}
Team Cohere and et~al.
\newblock Command a: An enterprise-ready large language model, 2025.

\bibitem{paulheim2018much}
Heiko Paulheim.
\newblock How much is a triple? estimating the cost of knowledge graph creation.
\newblock 2018.

\bibitem{dogruoz-etal-2021-survey}
A.~Seza Do{\u{g}}ru{\"o}z, Sunayana Sitaram, Barbara~E. Bullock, and Almeida~Jacqueline Toribio.
\newblock A survey of code-switching: Linguistic and social perspectives for language technologies.
\newblock In Chengqing Zong, Fei Xia, Wenjie Li, and Roberto Navigli, editors, {\em Proceedings of the 59th Annual Meeting of the Association for Computational Linguistics and the 11th International Joint Conference on Natural Language Processing (Volume 1: Long Papers)}, pages 1654--1666, Online, August 2021. Association for Computational Linguistics.

\bibitem{winata-etal-2021-multilingual}
Genta~Indra Winata, Samuel Cahyawijaya, Zihan Liu, Zhaojiang Lin, Andrea Madotto, and Pascale Fung.
\newblock Are multilingual models effective in code-switching?
\newblock In Thamar Solorio, Shuguang Chen, Alan~W. Black, Mona Diab, Sunayana Sitaram, Victor Soto, Emre Yilmaz, and Anirudh Srinivasan, editors, {\em Proceedings of the Fifth Workshop on Computational Approaches to Linguistic Code-Switching}, pages 142--153, Online, June 2021. Association for Computational Linguistics.

\bibitem{zhang2023speakforeignlanguagesvoice}
Ziqiang Zhang, Long Zhou, Chengyi Wang, Sanyuan Chen, Yu~Wu, Shujie Liu, Zhuo Chen, Yanqing Liu, Huaming Wang, Jinyu Li, Lei He, Sheng Zhao, and Furu Wei.
\newblock Speak foreign languages with your own voice: Cross-lingual neural codec language modeling, 2023.

\bibitem{huang2024zeroresourcecodeswitchedspeech}
Kuan-Po Huang, Chih-Kai Yang, Yu-Kuan Fu, Ewan Dunbar, and Hung yi~Lee.
\newblock Zero resource code-switched speech benchmark using speech utterance pairs for multiple spoken languages, 2024.

\bibitem{Ji_2022}
Shaoxiong Ji, Shirui Pan, Erik Cambria, Pekka Marttinen, and Philip~S. Yu.
\newblock A survey on knowledge graphs: Representation, acquisition, and applications.
\newblock {\em IEEE Transactions on Neural Networks and Learning Systems}, 33(2):494–514, February 2022.

\bibitem{zhou2021prix}
Wenxuan Zhou, Fangyu Liu, Ivan Vuli{\'c}, Nigel Collier, and Muhao Chen.
\newblock Prix-lm: Pretraining for multilingual knowledge base construction.
\newblock {\em arXiv preprint arXiv:2110.08443}, 2021.

\bibitem{Wang2022TextEB}
Liang Wang, Nan Yang, Xiaolong Huang, Binxing Jiao, Linjun Yang, Daxin Jiang, Rangan Majumder, and Furu Wei.
\newblock Text embeddings by weakly-supervised contrastive pre-training.
\newblock {\em ArXiv}, abs/2212.03533, 2022.

\bibitem{Li2023TowardsGT}
Zehan Li, Xin Zhang, Yanzhao Zhang, Dingkun Long, Pengjun Xie, and Meishan Zhang.
\newblock Towards general text embeddings with multi-stage contrastive learning.
\newblock {\em ArXiv}, abs/2308.03281, 2023.

\bibitem{nussbaum2025nomicembedtrainingreproducible}
Zach Nussbaum, John~X. Morris, Brandon Duderstadt, and Andriy Mulyar.
\newblock Nomic embed: Training a reproducible long context text embedder, 2025.

\bibitem{merrick2024arcticembedscalableefficientaccurate}
Luke Merrick, Danmei Xu, Gaurav Nuti, and Daniel Campos.
\newblock Arctic-embed: Scalable, efficient, and accurate text embedding models, 2024.

\bibitem{wang2024multilinguale5textembeddings}
Liang Wang, Nan Yang, Xiaolong Huang, Linjun Yang, Rangan Majumder, and Furu Wei.
\newblock Multilingual e5 text embeddings: A technical report, 2024.

\bibitem{chen2024bgem3embeddingmultilingualmultifunctionality}
Jianlv Chen, Shitao Xiao, Peitian Zhang, Kun Luo, Defu Lian, and Zheng Liu.
\newblock Bge m3-embedding: Multi-lingual, multi-functionality, multi-granularity text embeddings through self-knowledge distillation, 2024.

\bibitem{zhang2024mgtegeneralizedlongcontexttext}
Xin Zhang, Yanzhao Zhang, Dingkun Long, Wen Xie, Ziqi Dai, Jialong Tang, Huan Lin, Baosong Yang, Pengjun Xie, Fei Huang, Meishan Zhang, Wenjie Li, and Min Zhang.
\newblock mgte: Generalized long-context text representation and reranking models for multilingual text retrieval, 2024.

\bibitem{degand-muller-2020-introduction}
Liesbeth Degand and Philippe Muller.
\newblock Introduction to the special issue on dialogue and dialogue systems.
\newblock {\em Traitement Automatique des Langues}, 61(3):7--15, 2020.

\bibitem{günther2023jinaembeddingsnovelset}
Michael Günther, Louis Milliken, Jonathan Geuter, Georgios Mastrapas, Bo~Wang, and Han Xiao.
\newblock Jina embeddings: A novel set of high-performance sentence embedding models, 2023.

\bibitem{wang-etal-2024-improving-text}
Liang Wang, Nan Yang, Xiaolong Huang, Linjun Yang, Rangan Majumder, and Furu Wei.
\newblock Improving text embeddings with large language models.
\newblock In Lun-Wei Ku, Andre Martins, and Vivek Srikumar, editors, {\em Proceedings of the 62nd Annual Meeting of the Association for Computational Linguistics (Volume 1: Long Papers)}, pages 11897--11916, Bangkok, Thailand, August 2024. Association for Computational Linguistics.

\bibitem{dai2022promptagatorfewshotdenseretrieval}
Zhuyun Dai, Vincent~Y. Zhao, Ji~Ma, Yi~Luan, Jianmo Ni, Jing Lu, Anton Bakalov, Kelvin Guu, Keith~B. Hall, and Ming-Wei Chang.
\newblock Promptagator: Few-shot dense retrieval from 8 examples, 2022.

\bibitem{qu2021rocketqaoptimizedtrainingapproach}
Yingqi Qu, Yuchen Ding, Jing Liu, Kai Liu, Ruiyang Ren, Wayne~Xin Zhao, Daxiang Dong, Hua Wu, and Haifeng Wang.
\newblock Rocketqa: An optimized training approach to dense passage retrieval for open-domain question answering, 2021.

\bibitem{moreira2025nvretrieverimprovingtextembedding}
Gabriel de~Souza P.~Moreira, Radek Osmulski, Mengyao Xu, Ronay Ak, Benedikt Schifferer, and Even Oldridge.
\newblock Nv-retriever: Improving text embedding models with effective hard-negative mining, 2025.

\bibitem{Oord2018RepresentationLW}
A{\"a}ron van~den Oord, Yazhe Li, and Oriol Vinyals.
\newblock Representation learning with contrastive predictive coding.
\newblock {\em ArXiv}, abs/1807.03748, 2018.

\end{thebibliography}

\newpage

\appendix
\section{Plots for Model Comparison}\label{oth-plots}
The following plots represent some other comparisons between the models we compared between.
\begin{figure*}[ht]
    \centering
    \begin{subfigure}[b]{0.49\textwidth}
        \centering
        \includegraphics[width=\textwidth]{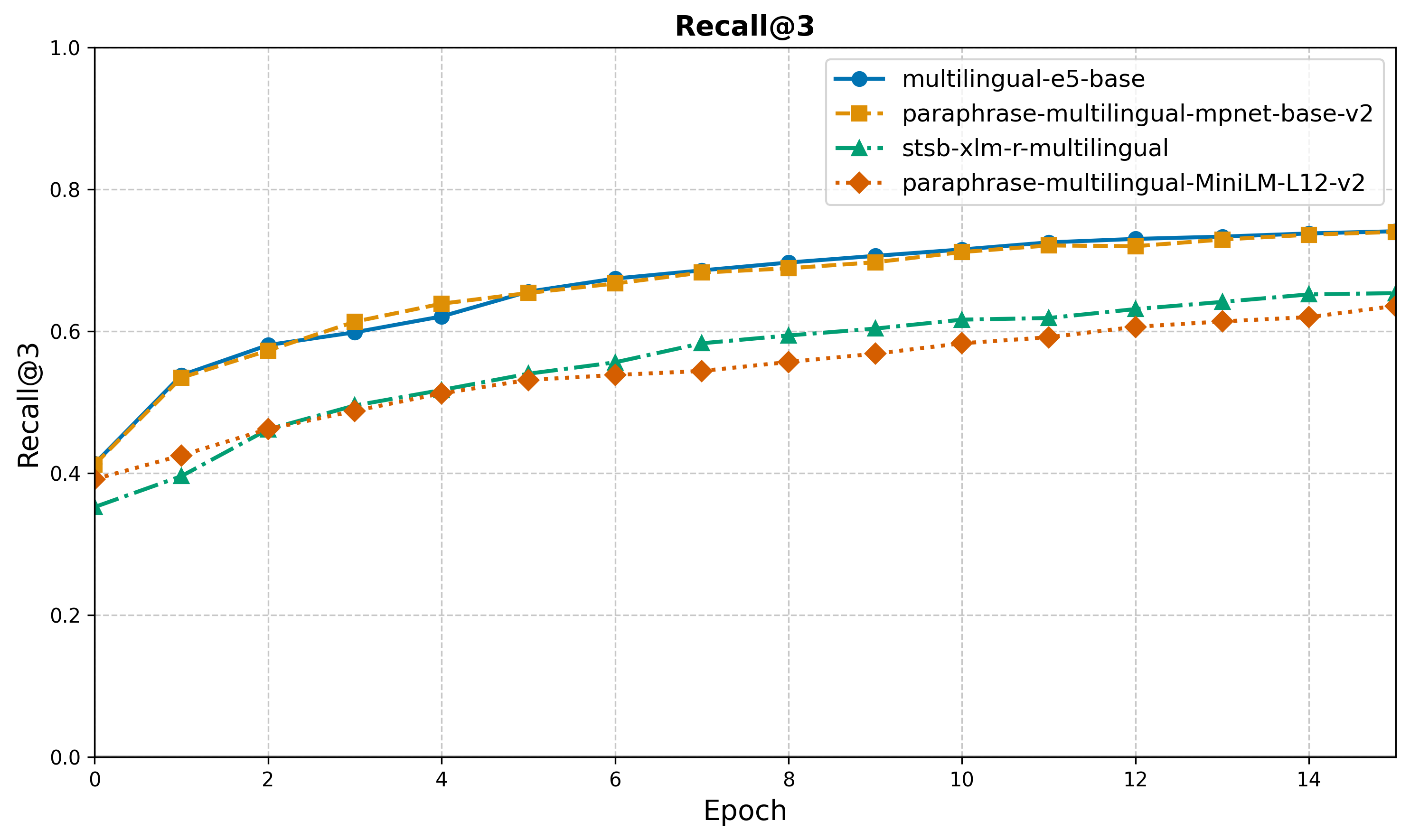}
        \caption{Performance of Recall@3}
        \label{fig:top3}
    \end{subfigure}
    \hfill
    \begin{subfigure}[b]{0.49\textwidth}
        \centering
        \includegraphics[width=\textwidth]{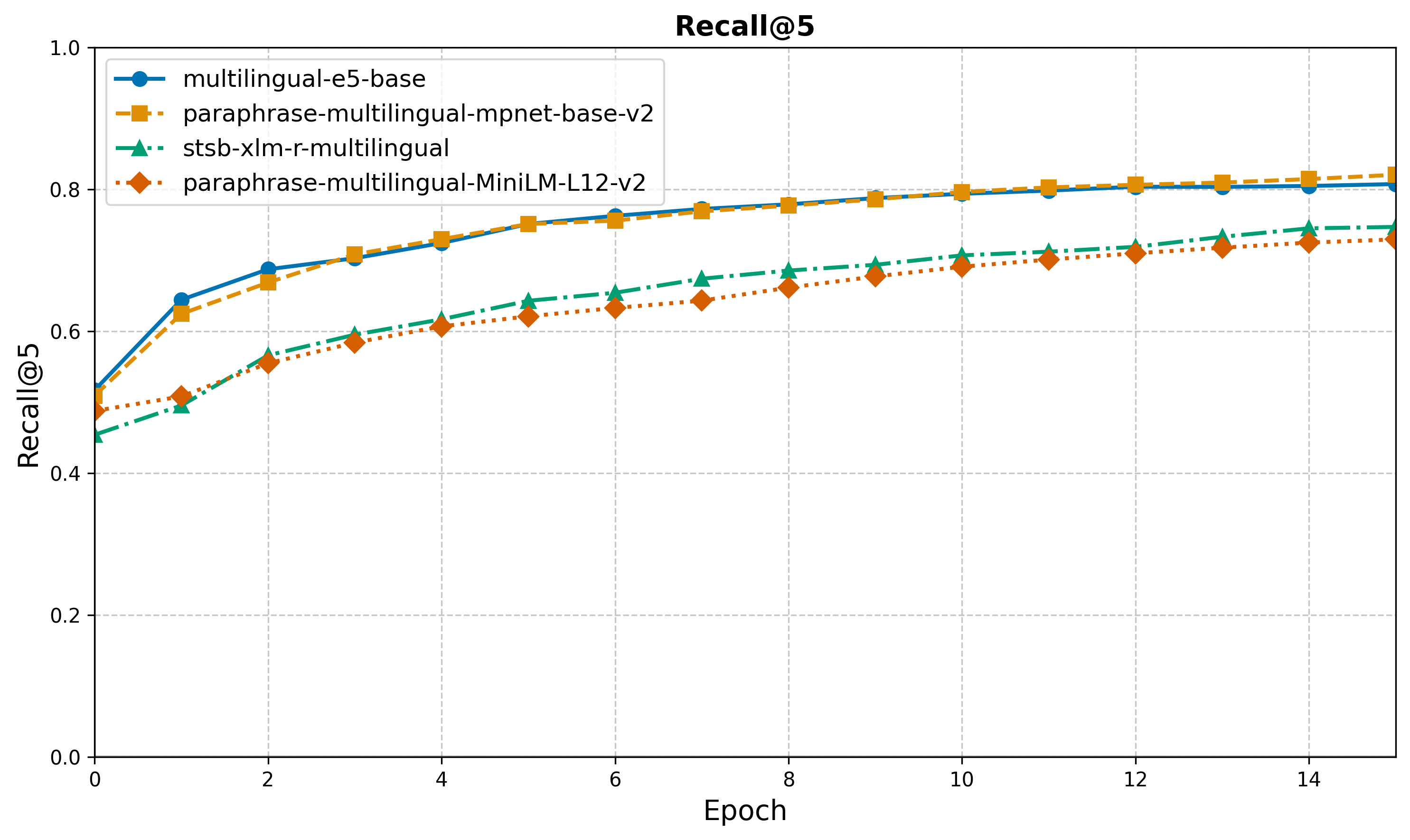}
        \caption{Performance of Recall@5}
        \label{fig:top5}
    \end{subfigure}
    \vskip\baselineskip
    \begin{subfigure}[b]{0.49\textwidth}
        \centering
        \includegraphics[width=\textwidth]{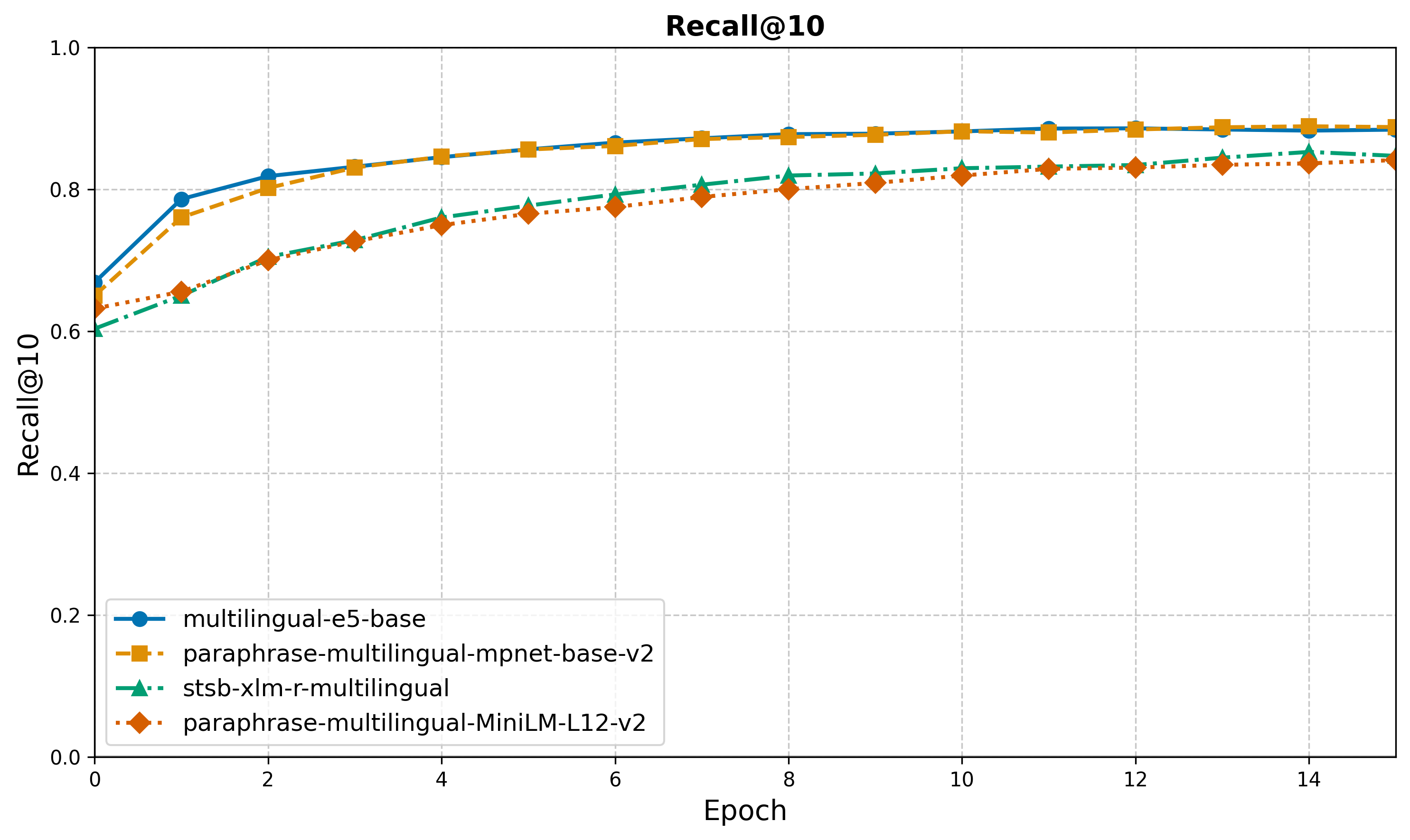}
        \caption{Performance of Recall@10}
        \label{fig:top10}
    \end{subfigure}
    \hfill
    \begin{subfigure}[b]{0.49\textwidth}
        \centering
        \includegraphics[width=\textwidth]{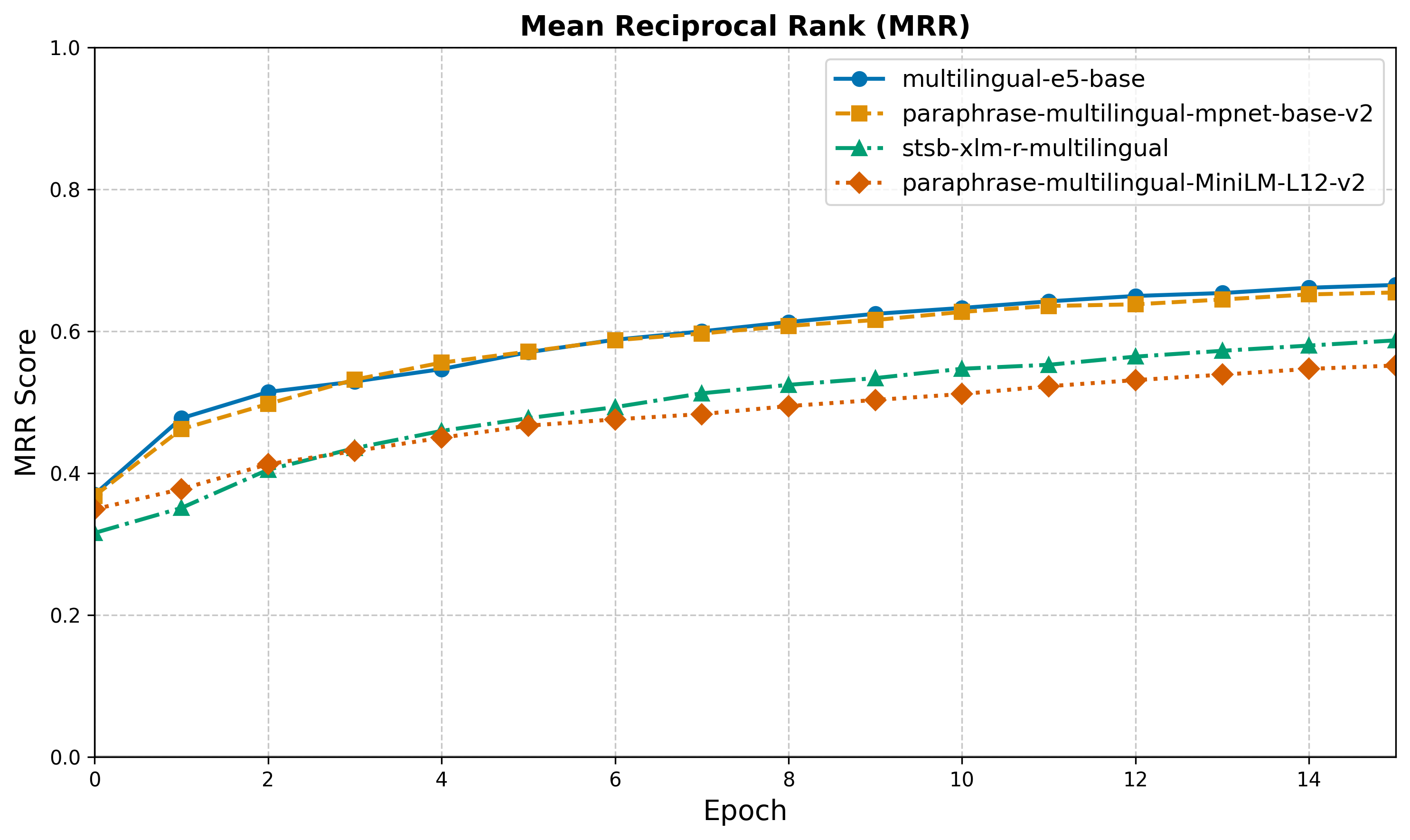}
        \caption{Mean Reciprocal Rank (MRR)}
        \label{fig:mrr}
    \end{subfigure}
    \caption{Performance metrics at different points during contrastive fine-tuning for various multilingual embedding models}
    \label{fig:all_metrics}
\end{figure*}

\end{document}